\documentclass[pdflatex,sn-mathphys-num]{sn-jnl}% Math and Physical Sciences Numbered Reference Style
\usepackage{graphicx}%
\usepackage{multirow}%
\usepackage{amsmath,amssymb,amsfonts}%
\usepackage{amsthm}%
\usepackage{mathrsfs}%
\usepackage[title]{appendix}%
\usepackage{xcolor}%
\usepackage[table]{xcolor}
\usepackage{textcomp}%
\usepackage{manyfoot}%
\usepackage{booktabs}%
\usepackage{algorithm}%
\usepackage{algorithmicx}%
\usepackage{algpseudocode}%
\usepackage{listings}%
\usepackage{caption}
\usepackage{subcaption}
\usepackage{listings} % highlight code
\theoremstyle{thmstyleone}%
\theoremstyle{thmstyletwo}%

\theoremstyle{thmstylethree}%

\begin{document}

\title[Article Title]{Sustainability assessment using multimodal AI agents}

\author*[1]{\fnm{Zhihan} \sur{Zhang}}\email{zzhihan@cs.washington.edu}

\author[1]{\fnm{Alexander} \sur{Metzger}}

\author[1]{\fnm{Yuxuan} \sur{Mei}}

\author[1]{\fnm{Felix} \sur{H{\"a}hnlein}}

\author[1]{\fnm{Zachary} \sur{Englhardt}}

\author[2]{\fnm{Tingyu} \sur{Cheng}}

\author[3]{\fnm{Gregory D.} \sur{Abowd}}

\author[1]{\fnm{Shwetak} \sur{Patel}}

\author*[1]{\fnm{Adriana} \sur{Schulz}}\email{adriana\_schulz@brown.edu}

\author*[1]{\fnm{Vikram} \sur{Iyer}}\email{vsiyer@uw.edu}

\affil[1]{\orgdiv{Paul G. Allen School of Computer Science \& Engineering}, \orgname{University of Washington}, \orgaddress{\city{Seattle}, \postcode{98195}, \state{WA}, \country{USA}}}

\affil[2]{\orgdiv{Computer Science and Engineering}, \orgname{University of Notre Dame}, \orgaddress{\city{Notre Dame}, \postcode{46556}, \state{IN}, \country{USA}}}

\affil[3]{\orgdiv{Electrical and Computer Engineering}, \orgname{Northeastern University}, \orgaddress{\city{Boston}, \postcode{02115}, \state{MA}, \country{USA}}}


\abstract{
Reducing the rapidly growing environmental impact of the computing industry requires assessing the emissions of electronics at scale. However, a traditional life cycle assessment (LCA) of an electronic device, which maps materials and processes to environmental impacts, often requires proprietary or unavailable data. Here, we reimagine conventional sustainability assessment by introducing a multimodal multi-agent AI system that emulates the collaborative process between LCA professionals and stakeholders (such as product managers and engineers) to automatically estimate the carbon footprint of electronic devices. The agents iteratively construct a complete life-cycle inventory by leveraging a structured data abstraction and software tools that mine information from the public internet, including repair communities and government regulatory databases. This reduces data gaps and data collection from weeks or months of expert time to under one minute. 
The system can calculate carbon footprint within 19\% of expert LCAs with zero proprietary data (typical of the variation between human LCAs). We also show that by encoding domain-specific knowledge, environmental impact estimation can be reframed as a data-driven prediction task, in which both unknown products and emission factors are represented as weighted combinations of similar ones with known emissions.\footnote{This 
article is published in \textit{Nature Electronics}, and is available online at: \url{https://www.nature.com/articles/s41928-026-01653-w}}
}

\maketitle
\section{Introduction}\label{sec:intro}

Information and computing technologies have a substantial and growing impact on the environment. Computing is estimated to account for 2.1-3.9\% of global greenhouse gas emissions~\cite{freitag_real_2021}, and is projected to surge to 20\% by 2030 without intervention~\cite{andrae_global_2015,wang_designing_2025}. There is also evidence of growing public awareness of these impacts: 70\% of consumers in a global survey were willing to pay a premium for more environmentally friendly electronics~\cite{kamiya2025rethinking, proserpio_impact_2024}, which is matched by rapidly increasing internet searches for product carbon footprints (Fig.~1a).
However, decarbonization of electronics is challenging. In addition to the energy consumption during the use-phase of a device, there are also embodied emissions from manufacturing, which can account for 50-82\% of the total lifecycle carbon footprint across device classes~\cite{wang_designing_2025,gupta_act_2022,dell_inc_pcf_2024}.
This creates a need to rapidly assess the emissions of electronic parts at scale both for designers to reduce embodied emissions~\cite{englhardt_incorporating_2025} and to communicate environmental impact (EI) to the public.

The electronics industry uses a robust framework of tools for analysing and simulating electrical metrics, such as power, timing, and frequency characteristics. Yet there are no equivalent tools for environmental metrics, such as carbon footprint.
The traditional approach to assessing EIs is life cycle assessment (LCA). A process-based LCA quantifies the resources used throughout the lifecycle of a product or service---from raw material extraction and manufacturing to end-of-life disposal---and maps them to environmental impacts~\cite{lee_life_2004}. This analysis is typically performed by dedicated LCA professionals who first construct a life cycle inventory (LCI) that identifies all the components, processes, and energy used throughout a supply chain (Fig.~1b). 
For electronics, this data is often missing, proprietary and not publicly available, or siloed across disparate sources even within the same organization, thus requiring slow and costly manual collection efforts coordinating multiple stakeholders~\cite{bamana_addressing_2021,tu_mitigating_2024,zhao_data-centric_2025, englhardt_incorporating_2025}. 
Moreover, inconsistencies in methodology, background data matching, data transparency~\cite{andrae_life_2010,zeng_recycling_2014,national_academies_of_sciences_engineering_and_medicine_current_2022,tu_mitigating_2024}, and regional variations in supply chains~\cite{tan_single-use_2023} create uncertain benchmarking and reproducibility challenges.
Next, life cycle impact assessment (LCIA), inventory entries are matched to emission factors in LCA databases or direct measurements, which convert grams of material or kilowatt-hours of energy to carbon emissions. When a database entry is unavailable, experts choose the closest match~\cite{balaji_flamingo_2023}. This labor-intensive workflow is challenging to scale, especially in sectors with complex electronics supply chains with hundreds of components in a single device~\cite{lu_ecoeda_2023, zhang_deltalca_2024, fairphone5Lca}.

In this Article, we report an AI-driven computational workflow for LCA. In particular, we develop a multi-agent system capable of autonomously generating a life cycle inventory for real-world products and estimating their impact (Fig. 1c). Our AI system leverages recent advancements in large language models (LLMs) ~\cite{openai_gpt-4_2023,team_gemini_2025}, vision-language models (VLMs)~\cite{liu_visual_2023}, and AI agents~\cite{tu_towards_2025,jones_whats_2025}. While domain-specific EI estimation tools have been reported~\cite{google-flights,google-maps-routing,autodeskPlugin,msSustainabilityCalc,makersite,sluicebox,imec,2030Calculator,chatty_ecosketch_2024,zhang_deltalca_2024}, they address a limited subset of the challenges in LCA, typically being focused to a specific use case, such as transportation~\cite{google-flights, google-maps-routing} and cloud computing~\cite{msSustainabilityCalc}, and require structured inventory inputs or proprietary design files, such as CAD~\cite{autodeskPlugin} or PCB layouts~\cite{zhang_deltalca_2024}. In contrast, our approach offers an end-to-end system that requires only a product name as input. We use the system to calculate the cradle-to-gate (that is, production; Extended Data Fig.~1) carbon emissions of electronic devices. 

To understand the practical challenges of scaling LCA, we conducted industry research and interviews with professionals~\cite{englhardt_incorporating_2025}. This led us to reframe LCA as a hierarchical information retrieval problem across a device’s supply chain. Two AI agents emulate the manual process whereby an LCA expert iteratively refines the LCI through consultation with various stakeholders representing different knowledge and expertise about the design, manufacture, and end-of-life for a product (Fig. 1c). By automating the distributed information-seeking process, this approach enables human LCA experts to perform rapid estimation and focus on analysis and methodology refinement. 

Our system makes three contributions that advance LCAs. First, we develop a series of multimodal information retrieval tools~\cite{ma_mms_2024} that enable our multi-agent AI system to automatically search previously untapped public data sources, such as volunteer repair communities and government agencies, thereby automating both high-level product attributes and detailed LCI data collection. This reduces the time spent collecting such data from weeks or months of expert time to under one minute and closes data availability gaps. Combined with standard LCIA using established emission factors, our end-to-end system outperforms state-of-the-art LLMs and VLMs on EI estimation and LCA-related tasks and can achieve estimates within 19\% of expert LCAs. We note that this is on par with human expert assessments, which due to methodological differences can have error margins ranging from 20\%~\cite{dell_inc_understanding_2023} to threefold~\cite{dell_inc_pcf_2024, zhang_deltalca_2024}.

Second, we show that by representing products through high-level features and domain-specific knowledge (such as technology node, memory capacity), EI estimation can be transformed into a learnable prediction problem. This enables direct estimation by mapping an input query to a weighted combination of products with similar attributes and known emissions, bypassing the need for constructing a fine-grained component-level inventory. This approach achieves a mean average percentage error (MAPE) of 12.28\% for desktop computers, displays, and laptops.
Applying the same principle, our agentic system can infer emission factors for inventory entries not covered by existing LCA databases.
In a human benchmarking study, in which domain experts selected the closest database match with assistance from search engines and LLMs, our approach improves MAPE from 143.87\% to 23.61\%.
Finally, we analyse the inference-time scaling of the agentic system and data scaling in life-cycle modelling, and discuss the implications of the proposed approach for future sustainability workflows.

\begin{figure}[t]
    \centering
    \includegraphics[width=\linewidth]{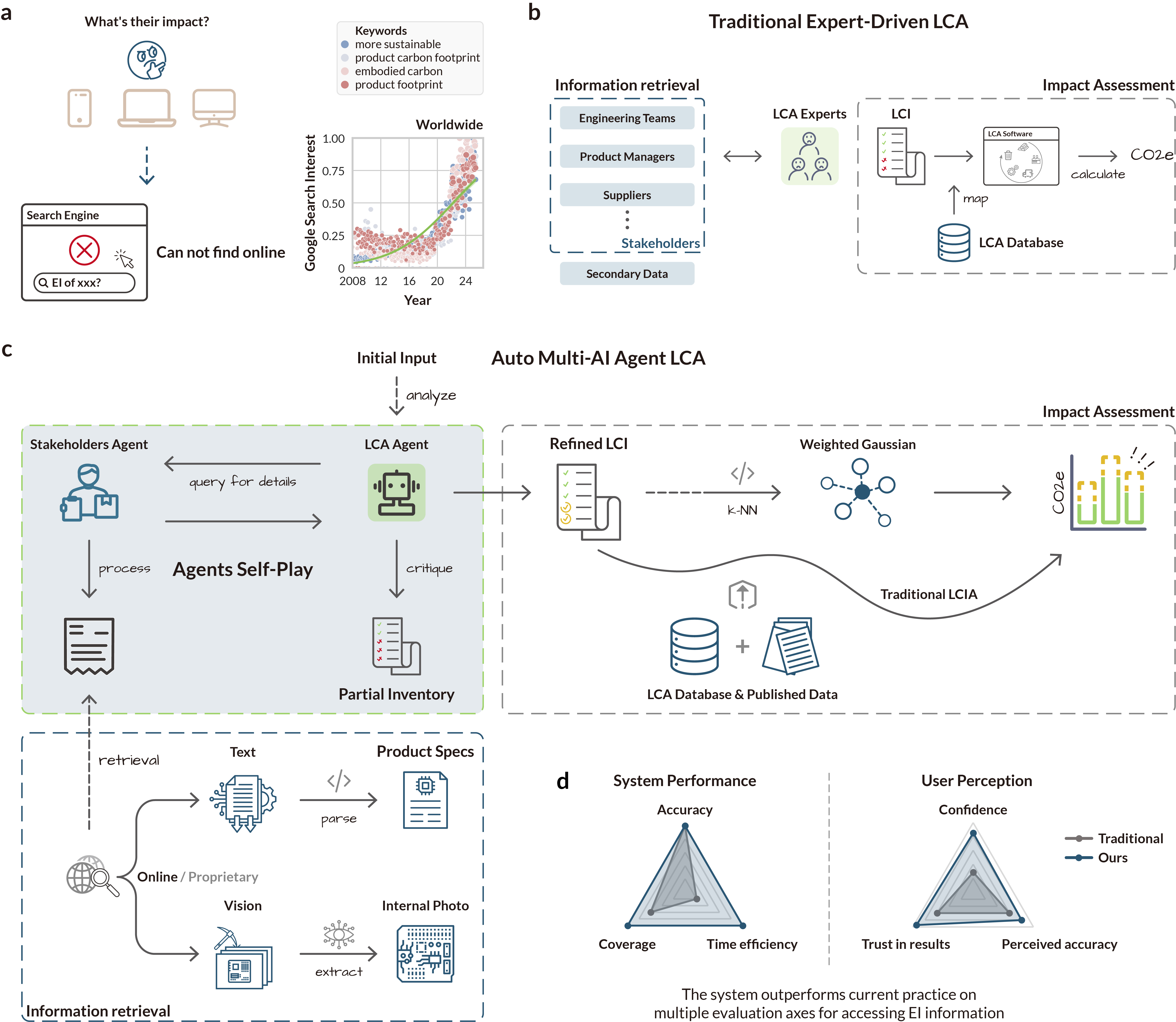}
    \caption{Autonomous life cycle assessment using multi-agent self-play.
    a, Google Search trends for sustainability-related keywords from 2008 to 2025, highlight growing public interest in considering EI in daily lives, however, such critical EI information remains largely unavailable especially for electronic devices. Search interest score is normalized monthly, with a value of 1 representing the peak popularity of the term. 
    b, Traditionally, EI is assessed via Life Cycle Assessment (LCA), a manual, expert-driven process. LCA experts construct a Life Cycle Inventory (LCI) by identifying all components involved throughout a product’s life cycle. This process requires slow and costly manual collection efforts, coordinating multiple stakeholders within the company and external suppliers, then mapping each entry in the LCI to an emission factor in databases. 
    c, We introduce the first autonomous multi-AI agent system capable of generating LCIs for real-world products and estimating their EI.
    The system simulates the traditional LCA process at scale through a multi-agent self-play environment in which an LCA expert iteratively refines the LCI by consulting with a variety of stakeholders representing different knowledge and expertise through iterative querying and dialogue. 
    The refined LCI can be used for standard LCIA to deliver final EI, or for estimation using a weighted sum of similar objects with known emissions based on domain-specific features.
    d, Evaluations show that the proposed system outperforms current practice in both technical performance (e.g., time efficiency), and user perception (e.g., confidence in accurately finding EI information) compared to conventional search approaches.
    }
    \label{fig:main_1}
\end{figure}

\clearpage
\section{Generating Inventory from Multimodal Information}\label{sec:LCI}
We develop and evaluate a multi-agent self-play system with automated feedback designed to emulate the core dynamics of traditional LCA expert workflows. Figure~1c shows the two stages of this workflow: 1) LCI generation, and 2) LCIA. Our system takes a product name as input and leverages the collaboration of LLM agents with specific instructions and tools~\cite{ma_mms_2024} to act as LCA experts and other stakeholders that iteratively work to retrieve LCI data from public data sources. These detailed inventory entries are multiplied by emission factors from an LCA database and summed following a conventional LCIA workflow. Additionally, we develop two alternative LCIA estimation techniques that improve this workflow. The first reduces the LCI detail needed by using a nearest-neighbor algorithm to directly estimate carbon from clusters of similar products. 
Second, we develop a data-driven approach to generate values for unknown emission factors to replace the need for expert heuristics. 

Generating a complete and accurate LCI is foundational for environmental impact assessment, yet remains one of the most labor-intensive stages in LCA. Traditional LCI construction demands significant proprietary supply chain disclosures such as a detailed bill of materials (BOM), domain expertise, and months of analysis~\cite{wernet_ecoinvent_2016}. This is particularly true for consumer electronics, which are complex assemblies of components, from integrated circuits to dense printed circuit boards with thousands of parts~\cite{fairphone4LCA, fairphone5Lca}. Due to the absence of standardized disclosures, even large companies often lack high-fidelity LCIs for their suppliers' products. We designed two key roles of AI agents that collaborate to autonomously generate LCIs from multimodal unstructured information, such as textual product specifications and visual teardown imagery:

\noindent\textbf{LCA Agent.} This agent acts as a critic. It creates data abstraction (DA) based on the query, which defines the relevant component classes to include in the inventory, ensuring the LCI is correct-by-construction with respect to system boundaries~\cite{zhang_living_2025}. It then assesses the LCI quality and completeness, identifies missing or ambiguous elements, and formulates targeted queries to the Stakeholders Agent to retrieve more information. This iterative exchange refines the LCI over multiple rounds.

\noindent\textbf{Stakeholders Agent.} This agent simulates the role of stakeholders such as engineers, product managers, and external suppliers, who contribute data. This agent leverages search tools to retrieve relevant information and also uses secondary data (e.g., research publications, industry reports) to supplement LCIs where direct access to proprietary or measured data is unavailable.

Our retrieval pipeline begins with the LCA agent constructing a DA (see Supplementary~3.1 for DA examples) from the user query, which can be a simple name (e.g., ``iPhone 12 Pro") or a photo. The DA serves as both a reasoning scaffold for agents and a constraint that bounds outputs within methodologically valid LCA representations~\cite{zhang_living_2025}. For electronics, this DA includes PCBs, ICs, sensors, passive electronic components (e.g., resistors, capacitors), and mechanical parts~\cite{zhang_deltalca_2024}. For devices with additional subsystems such as batteries or displays, the DA also guides the agent to retrieve relevant specifications, such as battery capacity and display type, respectively.

The Stakeholders Agent (Fig.~2a) searches for product specifications (see Methods) and teardown images (e.g., FCC reports, iFixit). It then extracts useful information by applying custom visual tools such as Fast Fourier Transform (FFT) filters for image preprocessing, and pre-trained deep learning models for object detection and segmentation. The agent performs dynamic multimodal reasoning by autonomously deciding which sources to use and the most appropriate tool for each subtask based on intermediate reasoning

We tested our image mining pipeline on a dataset of Apple products (see Methods) and achieved a 100\% success rate corresponding to the product name. For the FCC, we only failed on two recently released iPhone models whose FCC reports are still confidential; in those cases, the agent successfully retrieved internal photos from alternative online sources such as iFixit. The average runtime of the image mining process is 8.9 s.

A key visual subtask is to extract internal PCB photos from the hundreds of images returned from the search. These photos contain essential information about ICs and passive components as well as the area of the PCB itself, all of which directly impact EI. These images are unstructured and contain various internal electronic parts, such as batteries, speakers, or PCBs with metal shielding cans.
A second challenge is selecting the optimal board-level view from candidate PCB images, which may include zoomed-in or partial views. In such cases, the latest VLMs struggle to identify the complete board view (Fig.~2b). Our agent employs a combination of FFT (Fig.~2c) and YOLO-based electronic component detector (see Methods) to identify the image with the densest high-frequency content and the highest component count.
This toolchain not only achieves higher accuracy than VLMs but also offers significant computational efficiency, processing each image in 28 ms on an Apple M4 laptop. By contrast, Microsoft's lightweight Phi-3.5-vision-instruct~\cite{abdin_phi-3_2024} required 33.43 s of runtime per model to extract the same images even after quantization to 4-bit.

A component-level inventory is then created by detecting and classifying electronic components on the PCB using the aforementioned electronic detector. To enrich this inventory, we also employ OCR to extract potential text or part numbers on component packages, and initiate branched retrieval pipelines to gather additional information, such as process technology, package type, and dimensions (see Methods).

The physical dimensions of both the components and the PCB are crucial for accurate EI assessment.
When inferring the physical dimensions from images, the agent can calibrate pixel-to-distance conversion by cross-referencing any visible reference components with known dimensions (e.g., Apple's U1 chip sourced from datasheets or a ruler) in the image. 
The dimension estimation method is evaluated against products with publicly available PCB dimensions, such as the FairPhone and iPhone models. Our method achieved an MAPE of 5.48\%. For example, FairPhone 4's PCB is 153.0 $\times$ 67.0 mm~\cite{fairphone4LCA}, and our estimate is accurate to within 1 mm. The error is due to camera angles and image distortion near the edges. For models with irregular PCB designs, such as an L-shaped board, our bounding box also has errors of a few pixels.
It is worth noting that all VLMs consistently underperformed on this task due to their limited spatial reasoning for estimating physical dimensions and sizes~\cite{chen_spatialvlm_2024} (Fig.~2b).

To calculate the total EI, we multiply LCI entries by their closest emission factors from LCA databases such as Ecoinvent~\cite{ecoinvent}. We evaluate the end-to-end retrieval pipeline by comparing the CO2e estimates from the generated LCI for iPhones against values reported in Apple’s environmental reports. With zero proprietary information, our system achieves an MAPE of 18.02\% (Fig.~2d). 
To characterize the sources of error, we conducted a qualitative analysis by ranking estimation deviations and manually reviewing the products. The three with the lowest and highest deviations are shown in Fig.~2e (see Supplementary~3.5.1 for discussion of the outlier). 

We then ask whether this performance holds across the broader electronics industry. Testing on seven new product categories not seen during development, including GPUs and motherboards (Fig.~2f) to wearables and beyond (Supplementary Table~1), our total PCF estimates remain within 5–19\% relative to reported values. Together, these results demonstrate that our system can successfully retrieve relevant LCI data from public sources to either directly estimate EI or assist experts' data collection workflows without requiring proprietary supply chain information.

\clearpage
\begin{figure}[t]
    \centering
    \includegraphics[width=\linewidth]{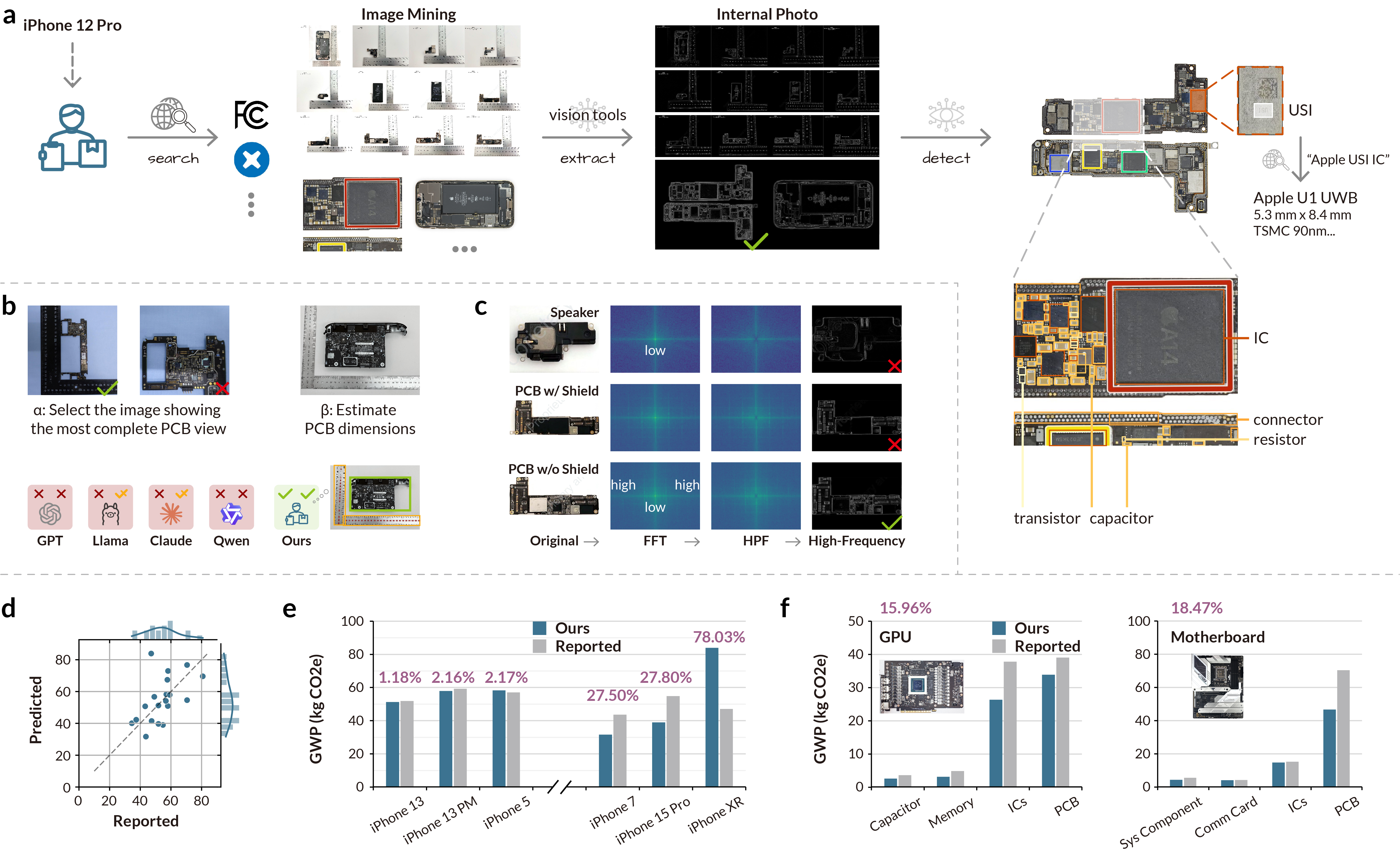}
    \caption{Multimodal agentic retrieval for life cycle inventory construction.
    a, Overview of the agentic multimodal retrieval pipeline. In this example, the agent begins by searching and parsing relevant images from online sources (e.g., FCC, iFixit). Internal PCB photos are extracted, and a component inventory, including component categories and dimensions, is generated using a set of visual tools, including FFT and YOLO for detection. Reference components with known dimensions (e.g., an Apple U1 UWB chip) are used to calibrate pixel-to-distance conversion.
    b, Example comparisons for two key vision tasks: extracting optimal internal photos and estimating dimensions, shown across leading VLMs including GPT-4o (gpt-4o-2024-11-20), Llama3.2-Vision (90B), Claude 3.5 Sonnet, and Qwen2-VL (72B). Orange checkmarks with a cross indicate plausible but incorrect responses.
    c, Illustration of using FFTs as a visual tool to identify PCB images. Images of iPhone 12 Pro from FCC and their corresponding FFT magnitude spectrums, shown before and after applying a Gaussian HPF with a cutoff at 32 cycles per pixel.
    d, Scatter plot comparing predicted and reported product carbon footprints for iPhone models using the agentic visual pipeline.
    e, Comparison between estimated and reported CO2e for iPhones, highlighting the top 3 products with the highest and lowest error rates.
    f, Comparison of estimated and reported CO2e for the GPU ROG Strix GeForce RTX 4090 and the motherboard ROG STRIX Z790-A, demonstrating the generalizability of the multimodal agentic retrieval pipeline.}
    \label{fig:main_2}
\end{figure}

\clearpage
\section{Estimating EI from Domain-Specific Textual Features}\label{sec:kNN}
The agentic LCI generation above automates traditional LCA workflows by constructing detailed, component-level inventories. Through this process, we observe that many complex products, especially consumer electronics, are assemblies of shared OEM components across different brands with similar production processes. For instance, Windows PCs typically use CPUs from Intel or AMD, and discrete graphics cards from NVIDIA, and many smartphones use Qualcomm's Snapdragon processor. This is due to the small number of companies with manufacturing facilities capable of producing these highly specialized components, and it is these ICs, such as processors and memory, that account for a substantial fraction of a device's EI~\cite{gupta_act_2022, arroyos_tale_2022}. This observation suggests that similarly configured products could provide a useful approximation.

We introduce an estimation method that bypasses the need for fine-grained LCI decomposition and instead approximates a product’s EI directly from high-level domain-specific textual features.
Our agent loop above automatically retrieves textual specifications from sources such as product pages, datasheets, and certification websites (see Methods). We curated a benchmark dataset of 1446 electronics products across major companies by parsing official product carbon footprint reports (Extended Data Fig.~2).
Each data entry pairs domain-specific features with the corresponding reported PCF. 
We note that reported GWPs vary significantly between companies due to differences in their system boundaries and other modeling assumptions (Fig.~3a).

To estimate the EI of an input object, we introduce a weighted k-nearest neighbors (k-NN) estimator with data availability-aware weighting schemes (see Supplementary~4) that explicitly handle the incomplete data typical of real-world products, providing both point estimates and interpretable uncertainty bounds. 
Figure~3b shows a t-distributed stochastic neighbor embedding (t-SNE) visualization of the feature space, revealing natural clusters of similar products. For a given query product, the estimator identifies $k$ nearest neighbors using Euclidean distance, assigns data availability-aware weights to each neighbor based on attribute completeness, and fits a Gaussian distribution to the neighbors' carbon footprints. The mean provides the point estimate, while the standard deviation captures epistemic uncertainty due to data sparsity, missing attributes, and variance among similar products.
Unlike black-box deep learning models, this estimator is explicitly designed to meet the explainability and traceability standards of LCA. Practitioners can inspect the retrieved nearest neighbors to verify their representativeness (Fig.~3b), and the distribution provides uncertainty bounds based on empirical data of what it takes to produce similar products.

We evaluate our estimator against publicly reported carbon footprints from these companies. Our model was trained on a dataset of Asus products, and we held out 20\% for testing and employed 5-fold cross-validation.
As shown in Fig.~3c-e, when tested on a subset of products from the same company, the model achieved an overall MAPE of 12.28\%. Performance by category is within 10-20\%, with MAPEs of 10.41\%, 19.77\%, 11.06\% for desktops, displays, and laptops, respectively. These results outperform the error margins reported in the official full LCA reports of technology companies, which typically range from 19\%~\cite{dell_inc_understanding_2023}.
Notably, all company-reported values fall within the model's 95\% confidence interval. 
Our weighted k-NN also demonstrates the efficiency on the training set size, which is important for the LCA domain where expert-validated data is limited. We evaluated its performance scaling for varying training dataset sizes. We randomly construct training datasets of sizes 5, 10, 20, 40, 80, and 120 using the Asus laptop data. As shown in Fig.~3f, model performance, measured in MAPE, improves as the training dataset size increases and converges at 80 samples.
Despite differences in LCA modeling methodologies across companies, our weighted k-NN generalizes well beyond its training distribution. When applied to Dell products using a model trained exclusively on Asus data, the estimator achieves a PCF MAPE of 16.54\% after distributional calibration (Fig.~3g-j), comparable to within-company performance. This suggests that existing PCF data within an industry can provide informative distributional bounds to rapidly estimate the carbon footprint of products from unseen manufacturers.

We further compared our model against GPT-4o~\cite{openai_gpt-4_2023} and three widely adopted ML regressions. Our model demonstrated superior performance across the board, ranking first in five of the six conditions and second-best in the remaining one (Supplementary Table~3).
Another key advantage of our method is that, compared to ML models, it does not require retraining to integrate new data. This is well-suited for LCA, as it has continuous updates of training data.

A within-subject user study (see Supplementary~6) shows that our system significantly improves user confidence, perceived accuracy, and trust in EI information compared to existing search tools, including Google Search and LLMs.

\clearpage
\begin{figure}[h]
    \centering
    \includegraphics[width=\linewidth]{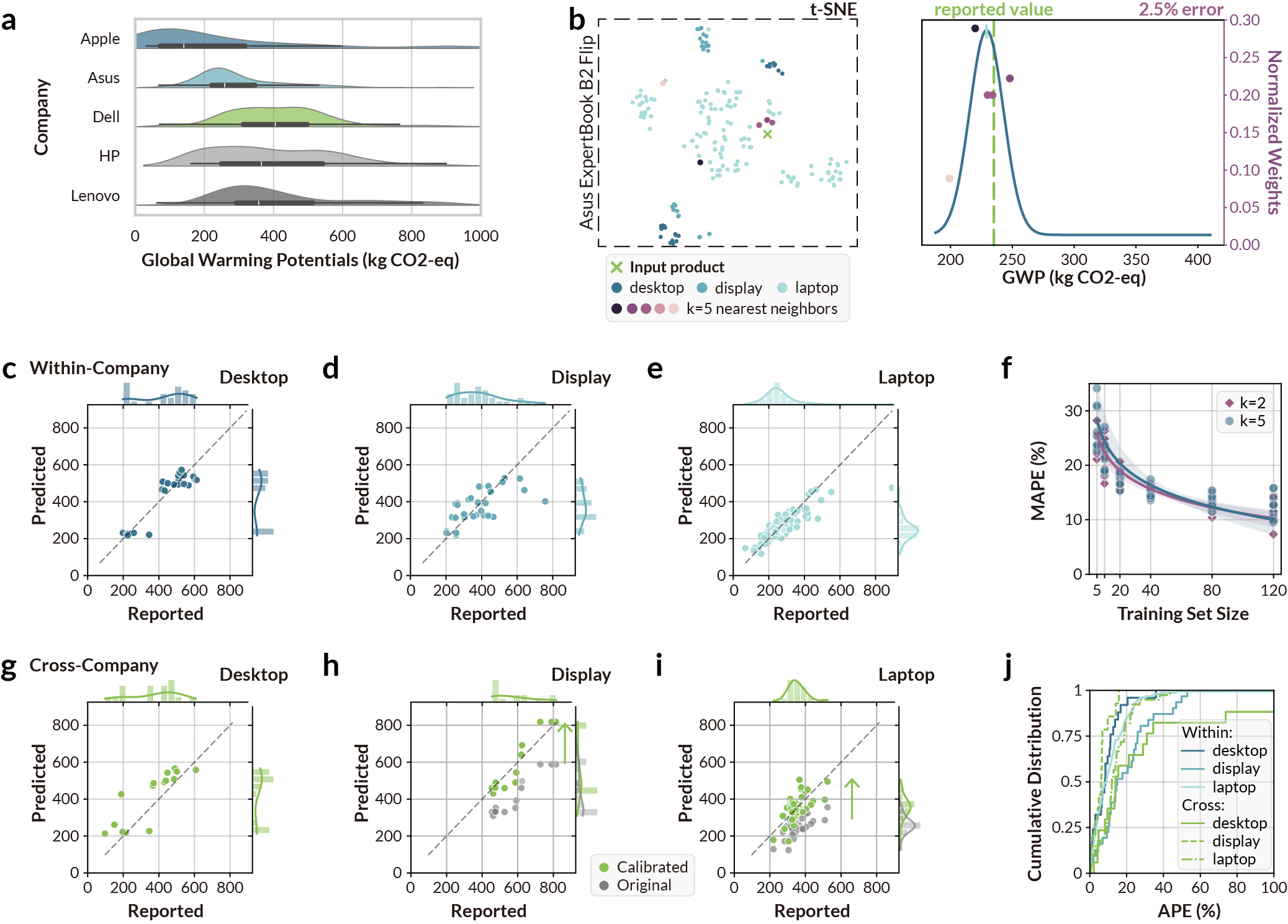}
    \caption{Environmental impact estimation from domain-specific textual features. 
    a, Distribution of reported product carbon footprints from environmental reports of global electronics companies, including Apple, Asus, Dell, HP, and Lenovo. Data are N = 1,229 product models; box plots show the median, 25th to 75th percentile, and $1.5\times$ interquartile range.
    b, Example inputs for weighted k-NN Gaussian carbon footprint estimation. The t-SNE plot (left) shows clusters of different product types and the closest data points based on Euclidean distance. Each data point represents a product model; the corresponding weighted Gaussian distributions over five nearest neighbors, weighted by the number of available attributes, with predicted and reported values highlighted (right).
    c-e, Scatter plots comparing predicted and reported product carbon footprints for within-company estimation across desktops (c), displays (d), and laptops (e); marginal distributions shown.
    f, Performance of our weighted k-NN as the training dataset size increases; data are N = 10 independent runs per dataset size, shaded region indicates 95\% CI of regression.
    g-i, Scatter plots comparing predicted and reported product carbon footprints for cross-company estimation across desktops (g), displays (h), and laptops (i); marginal distributions shown. Calibration based on overall distributional differences, which reflect variations in LCA methodologies across companies, improves cross-company estimation.
    j, Empirical cumulative distribution function (ECDF) of MAPE comparing across product categories between within- and cross-company.}
    \label{fig:main_3}
\end{figure}

\clearpage
\section{Generalizing Emission Factors}\label{sec:EF}
LCIA is typically straightforward: LCI entries are multiplied by their corresponding emission factors and summed to compute a total. However, this becomes challenging when LCI entries lack direct matches in LCA databases. Practitioners are then left with the labor-intensive or infeasible choice of manually measuring the process, or in many cases, using their expert judgment to select appropriate proxies. The latter approach is common but introduces subjectivity and increases uncertainty ranges. While recent works have explored automating emission factor mapping~\cite{balaji_flamingo_2023, balaji_caml_2023, balaji_emission_2025}, they remain limited to discrete entries in LCA databases that may not be representative.

To go beyond expert-driven heuristics, we instead develop a data-driven method to generate explainable estimates of unknown emission factors. First, our multi-agent retrieval framework autonomously derives class-specific features representing the LCI entry. Next, we apply k-NN clustering to identify the most representative known entries in the LCA database, followed by a weighted k-NN estimator to generate the final output. 

We evaluate our method on two representative LCA scenarios: carbon intensity estimation for electricity grids and for raw material classes such as plastics.
We compile a dataset of daily carbon intensities for 2024 across regional electricity grids in collaboration with Electricity Maps~\cite{electricitymaps}. The annual averages (Fig.~4a) reveal that geographic proximity is not a reliable predictor of carbon intensity. Neighboring regions often exhibit vastly different electricity profiles and emissions due to policy, infrastructure, and underlying geological environment.
To address this, we construct a feature-based representation of each grid's power profile, encoding the share of nuclear, wind, hydro, and other sources. The feature embeddings display clear gradients aligned with carbon intensity (Fig.~4b).
As shown in Fig.~4c, our method achieves effective performance even with limited data, reaching near-saturation with only 120 regional entries. This data efficiency is critical for generalizing to other LCA classes where labeled data is scarce. 
Using all 348 regional data points, our model achieves a coefficient of determination of 0.89, closely aligning with reported values (Fig.~4d).
The approach remains robust under missing data conditions (Fig.~4e), even when approximately 50\% of all feature entries are absent across the dataset, a frequent issue in real-world settings where information is often proprietary or not publicly available.

Next, we apply the same methodology to estimate CO2e for raw material entries in LCA databases. Traditionally, LCA professionals select the closest database entry using textual descriptions.
As shown in Fig.~4f, embeddings of textual descriptions for raw materials cluster at the high-level ISIC classes (e.g., wood, metals, petroleum). However, intra-class variance in carbon intensity is poorly captured by language-based representations alone (Fig.~4g). We find that incorporating class-specific features (see Methods), such as melting point, phase at standard temperature and pressure, and elemental category, markedly improves estimation accuracy (Fig.~4h for plastics and Extended Data Fig.~3 for metals and petroleum), underscoring the importance of encoding domain-specific knowledge into LCA modeling.

Finally, we benchmark our method against nine human experts in LCA and materials science (Fig.~4i), who are asked to select the closest emission factor match for masked raw material entries with assistance from standard tools, including search engines and LLMs. Our agentic estimation outperforms all human experts in both MAPE and MAE, even after excluding outliers (Fig.~4j,k). 
To assess the reliability of human expert annotations, we measured inter-annotator agreement across multiple metrics (see Supplementary~5.1). The intraclass correlation coefficient~\cite{koo_guideline_2016} (ICC(2,1) = 0.85) and Krippendorff's alpha (0.51) both indicate agreement, and semantic similarity analysis revealed high pairwise consistency among experts (0.81) and alignment with target products (0.76), confirming that experts make conceptually coherent selections. 
Yet semantic coherence does not guarantee numerical accuracy, for example, when presented with \textit{polystyrene, extruded} (EF = 10.5 kg CO2e kg$^{-1}$, ecoinvent v3.11), experts frequently selected \textit{polystyrene, general purpose} (EF = 3.75 kg CO2e kg$^{-1}$, ecoinvent v3.11). Despite near-identical nomenclature and shared polymer identity, the two entries differ in manufacturing stage: the extruded variant includes additional processing energy and associated emissions, resulting in a nearly threefold difference in carbon intensity (see Supplementary Table~4 for more examples).
Consistent with this pattern, the coefficient of variation across targets remained high at 46.9\%, underscoring the fundamental challenge in LCA practice: even when experts demonstrate strong agreement in distinguishing database entries, their numerical EI estimates can diverge substantially.

\clearpage
\begin{figure}[h]
    \centering
    \includegraphics[width=\linewidth]{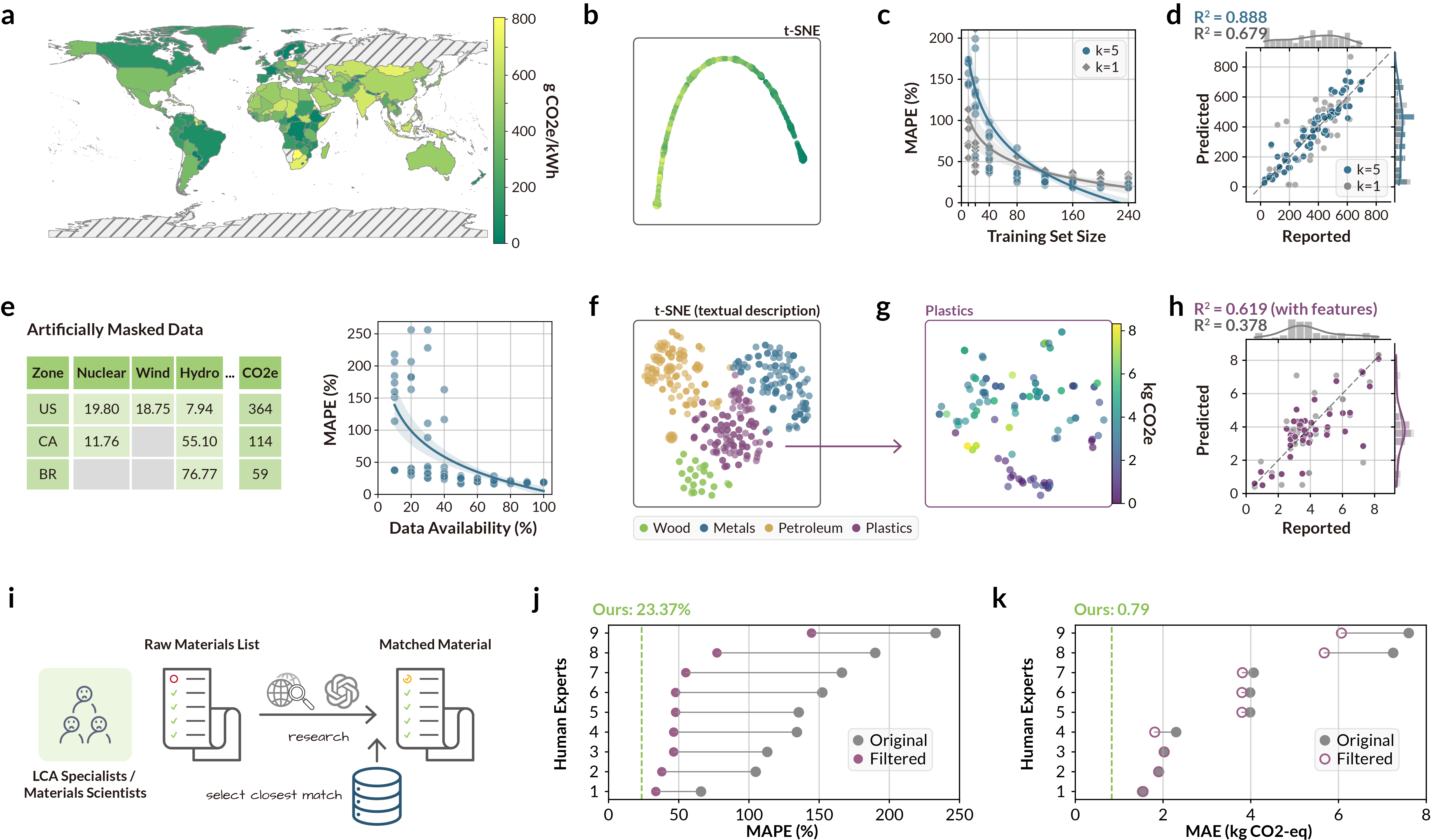}
    \caption{Emission factor estimation for life cycle impact assessment. 
    a, Global mean electricity carbon intensity by country in 2024, calculated from daily regional data from the collaboration with Electricity Maps~\cite{electricitymaps}.
    b, t-SNE embeddings for electricity power source features, colored by corresponding carbon intensity. Each point represents a geographic region.
    c, Performance of our weighted k-NN Gaussian electricity carbon intensity estimation as the dataset size increases, comparing different numbers of nearest neighbors used; data are N = 10 independent runs per dataset size, shaded region indicates 95\% CI of regression.
    d, Scatter plot comparing predicted and electricity carbon intensities from Electricity Maps; marginal distributions shown.
    e, Performance of estimation improves as data availability increases (i.e., fewer missing information, such as unavailable power source data); data are N = 10 independent runs per availability level, shaded region indicates 95\% CI of regression.
    f,g, t-SNE plot of raw material entries from ecoinvent based on embeddings of their textual descriptions, showing clusters by ISIC classes (wood, metals, petroleum, plastics) (f). Zoom-in on plastics with points colored by carbon intensity (g). Each data point represents one LCA database entry.
    h, Scatter plot of predicted and carbon intensities from ecoinvent for plastics, using textual description embeddings alone or combined with domain-specific features (e.g., melting point for plastics); marginal distributions shown.
    i-k, Comparison of our agentic estimation to nine human experts across 90 LCA entries each. To simulate current practices for handling missing entries in LCA databases, human experts in LCA and/or material science are asked to select the closest match for a masked raw material from the remaining database. They are assisted by tools, including Google Search and ChatGPT.
    Our method outperforms all human experts in both MAPE (j) and MAE (k), even after removing outliers (± 3 s.d., representing highly inaccurate selections).}
    \label{fig:main_4}
\end{figure}

\clearpage
\section{Scaling of AI Agents for LCA}\label{sec:scaling}
The empirical success of training-time scaling laws~\cite{bahri_explaining_2024} in deep neural networks, where performance improves with increased data and compute, has been widely demonstrated across domains such as language and vision~\cite{kaplan_scaling_2020}. Designing an AI-based LCA workflow to evaluate consumer products at scale must therefore consider the trade-off of accuracy versus computation, which incurs both monetary and environmental costs. To minimize these costs, we design our system using existing LLM backbones to avoid training or fine-tuning large foundation models, and we develop lightweight tools such as estimators that require only 100 data points. In this agentic architecture, we instead explore scaling performance by extending inference-time search and reasoning.

The inference-time scaling of AI agent is evaluated on three core dimensions: 1) thinking time, defined as computational time allocated for the agent to reason and interact with its tools; 2) reasoning steps, measured as the number of self-play rounds; and 3) retrieval breadth, quantified by the number of external documents accessed. Evaluation is benchmarked using regional power source compositions from Electricity Maps (See Supplementary~2 for full metric definitions and procedures).

As shown in Fig.~5a, increasing thinking time enables deeper multi-agent deliberation and broader retrieval, as reflected in a greater number of reasoning steps and documents read, at the cost of increased token usage (Extended Data Fig.~4).
Figure~5b-e show that longer thinking time reduces absolute percentage error in the final CO2e estimation and consistently improves task performance, as measured by L1 error, F1 score, and Jensen–Shannon divergence (JSD) on LCI generation. Across all metrics, we observe diminishing returns beyond 40 seconds. 
Figure~5f-l show similar improvement trends for both additional documents read and increased reasoning steps. 

\clearpage
\begin{figure}[t]
    \centering
    \includegraphics[width=\linewidth]{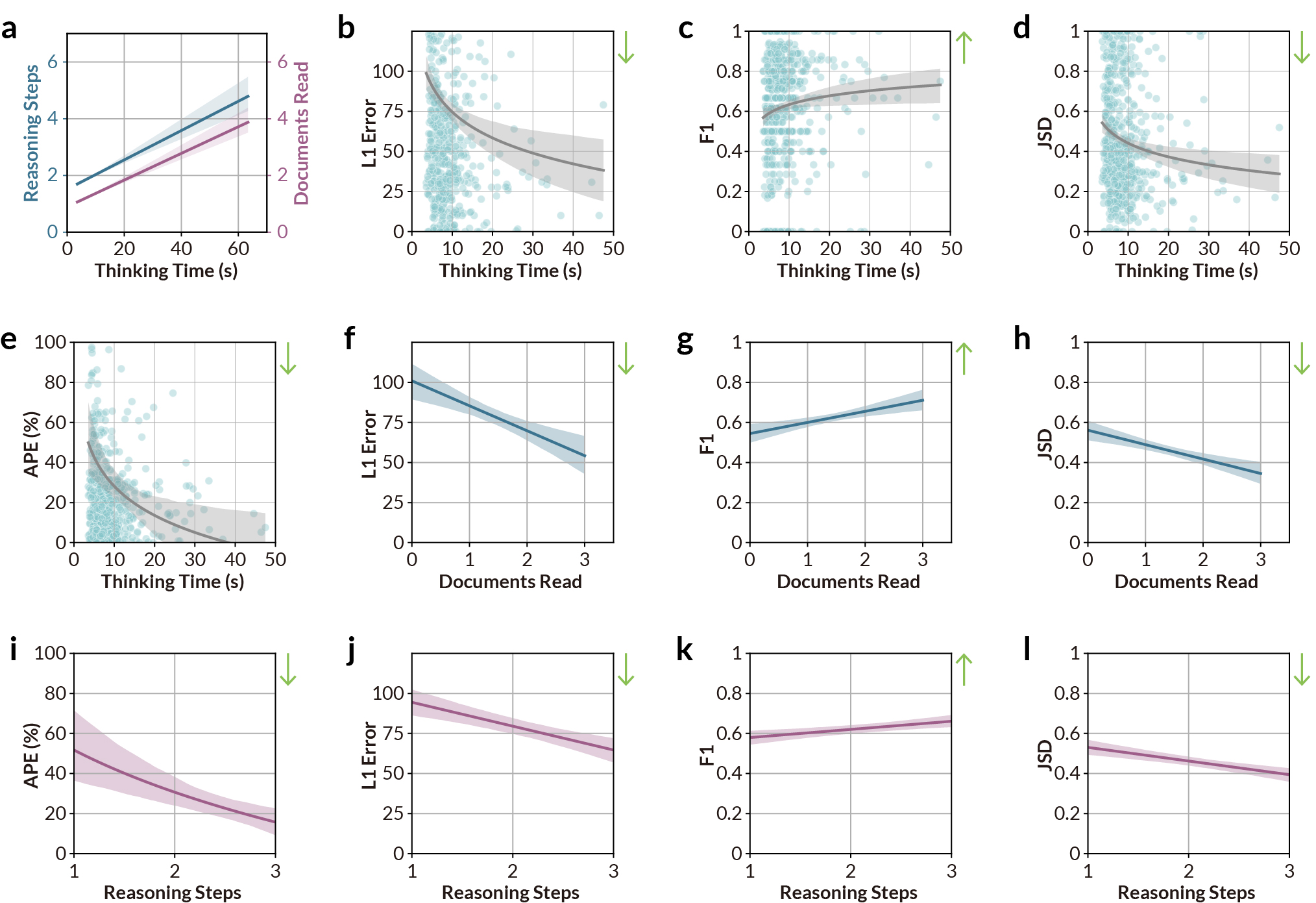}
    \caption{Inference-time scaling of agentic retrieval for life cycle assessment. 
    a, Number of reasoning steps and documents read increases with thinking time during agent self-play.
    b-i, Experiments show scaling thinking time (b-e), document read (f-h), and reasoning steps (j-l) are all effective on the life cycle inventory tasks through retrieval, as measured by L1 error, F1 score, and Jensen-Shannon divergence (JSD). Longer thinking times (e) and more reasoning steps (i) lead to lower absolute percentage error (APE) in final CO2e estimation.
    All data are N = 200 independent agent runs; shaded regions indicate 95\% CI of regression.}
    \label{fig:main_5}
\end{figure}

\clearpage
\section{Conclusions}
We have reported an AI-driven computational workflow to accelerate life cycle assessment. We demonstrated a multi-agent AI system that emulates human experts to automatically retrieve LCI data from public sources, a method to directly estimate EI based on clusters of similar products, and a data-driven approach to generate unknown emission factors. Each of these methods is shown to perform within 20\% of expert analyses or product carbon footprints with zero proprietary information.

These evaluations illustrate the promise of these methods but also raise important issues about the future use of automated tools. Unlike other domains, there is no unambiguously known ``true" value for an LCA or product carbon footprint: the analysis will reflect choices about the system boundaries, such as the granularity of parts and supply chains, and other modelling assumptions. Our agentic workflow therefore seeks to empower LCA practitioners to answer questions in multiple ways. 
First, automating data collection saves time, and using public sources to fill data availability gaps about subcomponents and emission factors could enable more comprehensive modelling. This frees experts to focus on methodological choices and quality of data sources, as well as enabling rapid iterative refinement of estimates. Second, neighbour-based estimation from clusters of similar products provides a useful first approximation when detailed LCI data are unavailable. Third, even when the absolute accuracy achieved is insufficient for formal carbon accounting, rapid estimates can be valuable for identifying environmental hotspots early in a product design cycle or the parts of a system most in need of detailed expert analysis. 

An important limitation of the neighbour-based estimator is that it characterizes the distribution of EI across existing, similar products, not the true impact of any specific product. The confidence intervals it produces reflect variability among reference products, and uses modelling assumptions that may be difficult to verify in practice. This is analogous to well-understood limitations in traditional LCA practice: ecoinvent~\cite{ecoinvent} proxy entries and industry averages describe a population of processes rather than a specific facility or supply chain. A manufacturer that has deliberately sourced renewable electricity or implemented novel low-emission processes would be expected to fall below any estimate derived from industry-representative neighbours. Nevertheless, we anticipate that this approach will provide LCA practitioners and product designers with practical means to obtain carbon footprint estimates and enable early-stage, informed decisions when exact data is unavailable.

As computational tools accelerate tasks like manual data collection for LCAs at scale, verification and human oversight become increasingly important. It remains an open challenge for AI systems to rigorously quantify the quality of agent-generated outputs. We implemented practical safeguards to collectively reduce errors but cannot eliminate them entirely. One advantage of our multi-agent architecture is that it makes certain decision points more transparent and modular. Ultimately, the LCA agent seeks to complement human LCA experts, enabling practitioners to leverage automated retrieval while retaining control over critical methodological choices. Validating agent outputs against companies’ internal supply chain data is an important direction for future work and would provide empirical reliability bounds. More broadly, optimizing the design of human–agent collaboration (e.g., determining when and how expert intervention is most valuable) remains an open question that could substantially improve the trustworthiness and efficiency of automated LCA workflows.

\clearpage
\section{Methods}\label{methods}

\subsection{AI Agents for LCA}
\subsubsection{Multi-agent architecture}
The multi-agent framework was developed using GPT-4o-mini as the LLM backbone, selected for its large context window of 128,000 tokens (at the time of development) and lower computational footprint relative to larger models. The framework was implemented using the OpenAI Agents SDK~\cite{OpenAI2025_AgentsSDK} and a custom Model Context Protocol (MCP) server that provides standardized tool access to both agents.

\paragraph{Agent roles and specialization}
Two specialized agents were instantiated with distinct roles in the LCA workflow: the LCA Agent and the Stakeholders Agent. The LCA Agent functions as a critical evaluator, responsible for designing DA, assessing inventory completeness and quality, and formulating targeted queries to address identified gaps or ambiguities. The Stakeholders Agent serves as an information retrieval specialist, utilizing available tools to gather data from external and internal sources in response to queries from the LCA Agent.

Each agent was equipped with role-specific system instructions incorporating chain-of-thought (CoT) prompting strategies tailored to their designated function. The LCA Agent receives instructions emphasizing methodological rigor, completeness criteria, and the ability to identify missing or inconsistent inventory entries. The Stakeholders Agent is instructed to maximize information retrieval using available tools, without explicitly specifying the tool sequence or priority.

\paragraph{Self-play protocol}
The self-play loop was structured as an iterative dialogue between the two agents. Each round begins with the LCA Agent issuing a critique or query based on the current state of the LCI. The Stakeholders Agent responds by executing tool calls over multiple reasoning steps to retrieve and extract relevant information. Retrieved data are then synthesized into a structured inventory update. The LCA Agent evaluates this response, updates the LCI, and determines whether additional information is required or if the inventory has reached convergence.

Convergence is defined as a state in which no critical data gaps remain, and the LCA Agent assesses the inventory as complete relative to the defined DA. If convergence is not achieved, the LCA Agent formulates a new critique that specifies the remaining deficiencies, and the cycle repeats. The self-play process terminates either upon convergence or after reaching a predefined maximum number of rounds.

\paragraph{Context management}
Conversation histories are maintained within each agent's context window throughout the self-play process. Messages generated by the agent itself are stored with the role designation ``assistant", while messages received from the counterpart agent are recorded as ``user". This bidirectional message structure enables each agent to maintain coherent dialogue state and build upon previous exchanges.
Upon completion of each input query, the context windows of both agents are reset.

\subsubsection{Tool orchestration}
The Stakeholders Agent has access to a suite of tools implemented through a custom MCP server. This extensible tool set was developed for electronics but can be expanded to address data retrieval challenges in other LCA domains~\cite{ding_scitoolagent_2025}. Representative tools include:

\textit{Textual retrieval.} The Google Custom Search API and OpenAI's built-in web search tools enable broad internet searches for product descriptions, specifications, and regulatory identifiers (e.g., FCC IDs). 
The URL-to-Markdown conversion tool processes web pages into structured text.

\textit{Visual retrieval.} The Google Custom Search API tool with image-specific parameters retrieves product photographs.
The FCC database scraper tool automatically extracts internal product photos from Federal Communications Commission certification databases using an FCC ID. 
The computer vision pipeline (detailed below) processes these images to extract component inventories and dimensions.

\textit{Emission factors.}
The Electricity Maps API provides regional electricity carbon intensity data. 
A geolocation tool converts between latitude and longitude coordinates and human-readable location names. 
An IP-based geolocation tool automatically identifies the user's location for region-specific use phase calculations. 

Each tool includes a natural language description that enables the Agent to select appropriate tools based on the current information need. The Stakeholders Agent was not instructed regarding tool selection order, priority, or conditional logic. Instead, tool descriptions enable the agent to dynamically determine appropriate tool sequences based on the LCA Agent's critique and inventory state. When a critique identifies missing component dimensions, for example, the agent may autonomously search for FCC photos or, alternatively, for teardown videos from repair communities. This adaptive tool orchestration emerges from the agent's reasoning capabilities rather than from predefined decision trees.

\subsection{Multimodal information retrieval}
Building on the agent architecture described above, the system retrieves LCI data from diverse heterogeneous sources. Relevant information is scattered across the public web in unstructured and multimodal formats, including: 1) product descriptions, specifications, and datasheets from manufacturers; 2) public disclosures required for regulatory agencies; 3) details and teardowns shared by online communities. 

\paragraph{Image mining}
The visual retrieval draws from two primary data sources:
\begin{itemize}
    \item \textbf{Government Agencies.} For example, the Federal Communications Commission (FCC) manages and licenses the electromagnetic spectrum, ensuring that devices do not cause harmful interference with radio communications. 
    As part of its regulatory responsibilities, the FCC maintains a database that includes detailed information about electronic devices with radio components such as Wi-Fi, Bluetooth, or other transmitters. 
    The FCC certification process requires the submission of internal photos, which are publicly available (We note that companies can request to make such information confidential. As a result, internal photos of some recent devices, such as Apple products released within the last two years, may be unavailable). 
    \item \textbf{Public Teardowns and Disassembly.} Online communities, such as YouTube, iFixit~\cite{iFixit}, TechPowerUp~\cite{TechPowerUp}, and LaptopMedia~\cite{LaptopMedia}, 
    provide detailed teardown photos and disassembly guides for various devices. These resources offer valuable clues for analyzing the components within a device.
\end{itemize}

The visual retrieval begins by issuing automated web and image searches using search tools like Google Image Search with various keywords such as ``teardowns", ``disassembly pictures", ``PCB analysis", while simultaneously locating the product's FCC ID through web scraping.
The FCC ID is a unique identifier assigned to electronic devices that have been certified by the FCC.
Given this ID, the agent can construct URLs to access relevant pages on FCC databases, navigate the HTML structure, extract rows labeled ``Internal Photos", and identify hyperlinks to download the associated PDF files, which often contain multiple pages with embedded images. The images are extracted using the PyMuPDF tool for further analysis.

\paragraph{Internal photo extraction} 
A two-stage filtering tool is developed to efficiently isolate relevant PCB images with low computational overhead.
The first stage leverages the insight that PCBs feature numerous tiny SMD components with sharp transitions between colors and fine textures, which correspond to high-frequency signals in the frequency domain. 
We proceduralize this insight by applying a FFT with a Gaussian high-pass filter (HPF) and Canny edge detector to identify images with the densest high-frequency content and edges as PCB candidates.

We also employ a custom-trained YOLO v11 model~\cite{redmon_you_2016} (see Supplementary~3.3) to detect PCBs and 32 component classes, including ICs, capacitors, resistors, transistors, and connectors. We set the confidence threshold to 0.3 and the Intersection over Union (IoU) threshold to 0.5, effectively reducing false negatives. We opted for a standard dense YOLO rather than larger transformer-based architectures~\cite{dosovitskiy_image_2020} because of its fast inference and low training cost. The images with high total component count are selected for inventory construction. Beyond filtering, the same YOLO model provides the component-level detections used in the following step.

\paragraph{Component inventory construction} 
To infer physical dimensions from imagery, a conversion ratio between pixel dimensions and real-world measurements is estimated using reference components whose dimensions are publicly available online.
We also observe that regulatory images, such as FCC submissions, typically include scale references such as a ruler. The scale marks are localized using an edge-detection tool, and the scale units (e.g., millimeters or inches) and their values are extracted using the Tesseract OCR library.
The pixel-to-distance ratio can also be determined by analyzing the range of these numbers and their corresponding positions on the image. 
We achieved a 100\% success rate for the ruler, number, and unit detections.

The SAM segmentation model~\cite{kirillov_segment_2023} is utilized to identify the PCB outline and generate a mask.
The derived pixel-to-distance ratio allows the agent to compute the dimensions of PCBs and each individual component.
Component category, physical dimensions, and other detailed semiconductor characteristics are subsequently matched to emission factors from prior literature~\cite{liu_future_2014, fairphone4LCA, gupta_act_2022, zhang_deltalca_2024} and LCA databases, and aggregated using standard LCIA methodologies to estimate the total EI.

\subsection{Dataset collection}
To evaluate the proposed autonomous LCA system, a benchmark LCA dataset was collected. In this work, we chose to focus on electronic products because information and computing technologies have a significant and growing impact on the environment. 
Electronic products also present a particularly challenging case for automated LCA due to their structural complexity. Unlike simple objects like plastic bottles, which consist of a single material for the body, modern electronic devices are assemblies of hundreds to thousands of mechanical and electronic parts (Extended Data Fig.~1b).

To ensure the correctness of retrieved information, such as reported carbon footprint values, researchers manually reviewed and validated each result before inclusion.
While this study uses company-reported product carbon footprint (PCF) values as the reference baseline, we note that these reports are not ground truth, due to a lack of transparency in underlying system boundaries and assumptions~\cite{andrae_life_2010}.

\paragraph{Product environmental reports}
Product environmental reports, which are a subset of LCA disclosures, typically present estimates of a product's total carbon footprint or GWP and a breakdown of those emissions across different life cycle stages, such as manufacturing, transport, and use (see example in Extended Data Fig.~2). Some companies also include additional details, such as the recyclability of materials, comparisons to prior models, and basic technical specifications.
In recent years, some large technology companies have dedicated resources to evaluating the GWP of their flagship products. 

The agent begins by attempting to locate the product's environmental report through an automated web search.
It formulates multiple search queries, combining the product name with different keywords such as ``Environmental Report", ``Carbon Footprint Report" and ``PCF", to increase recall. It uses the Google Custom Search API and retrieves the top two results for each query. 

If an environmental report is located, typically in PDF format, the agent employs the PyPDF and OCR Tesseract libraries to extract the document's content
The extracted content is then summarized using an LLM to extract essential details with the following prompt: ``Given the extracted text content from a product environmental report, extract the following key details:"

For companies whose environmental reports follow consistent formats (e.g., Apple, Dell, Asus), company-specific parsing tools were developed in Python for the agent to optimize processing time and reduce computing costs. These tools programmatically parse PDF reports for key product attributes, such as total carbon footprint and distribution, product weight, and lifetime, by applying predefined regular expressions tailored to each company's document structure.

\subsection{Domain-specific feature engineering}
Although textual descriptions of products and materials are readily available, generic text embeddings alone fail to capture the intra-class variance in EI, we therefore engineer domain-specific features grounded in manufacturing and materials knowledge for both electronics products and raw materials.

\subsubsection{Electronics products}
Drawing on research in electronics manufacturing, a set of attributes was hypothesized to contribute to the EI of electronic devices, including:
\begin{itemize}
    \item \textbf{Integrated Circuits.} ICs, including CPUs and GPUs, are evaluated based on their generation and series. Newer generations often utilize advanced, smaller technology nodes, which typically result in lower yields and increased energy consumption during fabrication~\cite{zhang_deltalca_2024}, thereby raising the environmental impact per chip~\cite{gupta_act_2022}. We note that chips within the same generation and series are often manufactured using identical designs and technologies, then categorized into different tiers through a binning process based on core quality~\cite{zolotov_voltage_2009}. Since chips of the same generation share the same die sizes and technology nodes, the two main indicators of IC's EI, they are considered equivalent.
    
    \item \textbf{Random-Access Memory.} RAM generations, such as DDR4 and DDR5, reflect successive technological advancements that offer higher density, thereby reducing the EI per unit byte~\cite{gupta_act_2022}. Higher RAM capacities, however, increase the environmental impact due to the additional materials and energy required for production.
    
    \item \textbf{Storage.} Storage devices, including HDDs and SSDs, have distinct environmental impacts~\cite{gupta_act_2022}. However, within each storage type, larger capacities correlate with higher EIs.
    
    \item \textbf{Batteries.} In portable devices like laptops, larger batteries generally result in higher EIs due to the increased use of toxic materials and energy during manufacturing.
    
    \item \textbf{Display.} Larger sizes and higher refresh rates demand more energy to produce and operate. We also categorize displays by panel type, because different panel types, such as IPS and OLED, also vary in EI due to distinct structures.
    
    \item \textbf{Physical Dimensions.} Larger physical volumes and weights tend to increase EIs due to higher material use and associated transportation emissions.
    
    \item \textbf{Energy Consumption.} Different modes of energy consumption, such as active mode and sleep mode, contribute variably to a product's EI. Higher power consumption in any operational mode correlates with increased environmental impact due to greater energy use over the lifetime.
\end{itemize}
40 features are then derived from the hypotheses for product specification retrieval to ensure diverse coverage.

The agent processes the first 10 search results to extract relevant specifications. 
Additionally, the agent searches the product in three popular electronic certification databases---Energy Star~\cite{EnergyStar}, EPEAT~\cite{EPEAT}, TCO~\cite{TCO}---for complementary information. 
For example, Energy Star provides detailed information on a product's energy consumption and efficiency; TCO Certified Edge requires displays to contain at least 85\% recycled plastic and be halogen-free; EPEAT-registered products have a score and tier system that indicates the number of optional environmental benefit criteria met. 
The entire process, including text embedding, takes less than 10 seconds per product. 

We note that many product series include numerous similar variants, i.e., different model numbers, such as the Dell OptiPlex 7000 desktops~\cite{dell_inc_pcf_2024}, where models typically differ only in minor attributes. 
To minimize overfitting and bias, the dataset was streamlined by randomly selecting one representative model from each product series. 
The finalized dataset sizes are summarized in Supplementary Table~2.

\subsubsection{Raw materials}
We derive 25 features to balance coverage diversity with computational efficiency. These features span physical properties, chemical hazards, and production context.

\begin{itemize}
    \item \textbf{Material Properties.} Fundamental physical descriptors such as molecular weight, density, melting point, solubility in water, and phase at standard conditions characterize how substances behave and transform under ambient or process-specific conditions. These attributes affect energy use during manufacturing, transport, and storage. For instance, materials with lower melting points generally require less thermal processing energy, thus lowering EI.
    
    \item \textbf{Mechanical Limits.} Properties like tensile strength and elasticity indicate the mechanical robustness of a material, which often correlates with the complexity of its synthesis. High-performance materials typically demand more controlled environments or energy-intensive pathways, indirectly raising their EI.
    
    \item \textbf{Process Efficiency and Intensity.} Features describing a material’s manufacturing temperature and pressure thresholds, overall process yield, and energy classification serve as proxies for the efficiency and intensity of production. 
    Indicators such as manufacturing yield, manufacturing temperature and pressure thresholds reflect the efficiency and intensity of production. Materials that require extreme conditions, or involve inefficient synthesis are typically associated with higher EI. 
    Additionally, the use of toxic or hazardous reagents indicates the potential post-processing mitigation or disposal steps, both of which are tightly coupled with EI.
    
    \item \textbf{Chemical Safety and Environmental Risk.} A set of binary hazard indicators, based on such as Globally Harmonized System (GHS) labeling, reflects the material’s toxicity, flammability, corrosiveness, reactivity, and environmental harm. Materials flagged under these categories often necessitate additional containment, purification, or disposal steps, which increase both direct and indirect EI.
    
    \item \textbf{Industrial Scale and Economic Context.} Real-world constraints such as global production volume and market price offer insights for the material's role in the industrial ecosystem. Common, Commoditized materials with large-scale production often benefit from optimized, lower-impact production methods, while niche or high-cost materials often rely on specialized, less efficient manufacturing pathways.
\end{itemize}

\subsection{Human expert benchmark}
Nine human domain experts with professional backgrounds in LCA and/or materials science were recruited via industry partnerships and snowball sampling.
The study utilized 116 unique emission factor entries originating from ecoinvent~\cite{ecoinvent} v3.11 database's ``market activity" entries within ISIC classes 2013 (Manufacture of plastics and synthetic rubber in primary forms) and 2220 (Manufacture of plastics products).
Each participant independently evaluated 90 target entries, presented across nine trials in which users were given a list of 10 targets each. For each list, participants selected the most appropriate emission factor match from a candidate pool of 106 emission factor entries (the remaining entries after excluding the 10 targets). 
Between trials, we rotated which entries served as targets, ensuring no repetitions while maintaining a consistent pool size of 106 candidates per sheet.
The task structure (10 targets per trial × 9 sheets) was designed to balance consistency with model evaluation dataset partitioning and participant workload, preventing fatigue.

All participants received a standardized task briefing that included background information, instructions for interpreting LCA documentation, and guidance on recording their selected match and the associated database link in designated fields.
To simulate real-world conditions, participants were allowed to consult external resources, such as Google Search and LLMs (e.g., asking ChatGPT, ``What is the potential closest match for [material] from the following candidates in an LCA database?"). However, participants were recommended to rely on their own judgment when making the final selection.

\clearpage
\section{Data availability}
All data needed to evaluate the conclusions of this study are available in the paper or in the Extended Data and Supplementary Information. 
Source data are provided with this paper.

\section{Code availability}
The source code is available for download on GitHub at https://github.com/iamZhihanZhang/AI-Agents-for-LCA. Future updates and new releases will also be available at this link.

\section{Acknowledgements}
We thank Electricity Maps for providing electricity carbon intensity data, S. Yu for feedback on the figures, and all anonymous participants for their contributions to the human studies of this work, which were approved by the University of Washington Institutional Review Board (STUDY00023836).
This research was supported by Amazon Research Awards, Alfred P. Sloan Foundation, and the National Science Foundation under award numbers CNS-2401177, CNS-2338736, CNS-2310515, and IIS-2212049.
Z. Zhang was supported by the Google PhD Fellowship.

\section{Author Contributions Statement} 
Z.Z., Y.M., F.H., Z.E., A.S. and V.I. conceptualized, organized and structured the work. 
Z.Z. and A.M. designed and developed the multi-agent system and retrieval pipeline, and conducted evaluations.
Z.Z., Y.M. and F.H. designed the estimation from textual features, experiments and evaluations.
Z.Z. designed the emission factor estimation experiments and evaluations.
Z.Z. and A.M. designed the agent scaling experiments and evaluations.
Z.Z. and T.C. designed and conducted the human studies. 
Z.Z., G.D.A. and V.I. wrote the manuscript. 
G.D.A., S.P., A.S. and V.I. jointly supervised the work. 
All authors contributed to the study concept and experimental methods, discussed the results and edited the manuscript.

\section{Competing Interests Statement}
S.P. is an employee of Google LLC. V.I. is an employee of Amazon.com. 
Z.Z., Y.M., F.H., Z.E., S.P., A.S., and V.I. are inventors on a US provisional patent application (63/853,996) submitted by the University of Washington, which is related to this work.
A.M., T.C., and G.D.A declare no competing interests.

\clearpage
\section{Extended data}

\begin{figure}[h]
    \centering
    \includegraphics[width=0.6\linewidth]{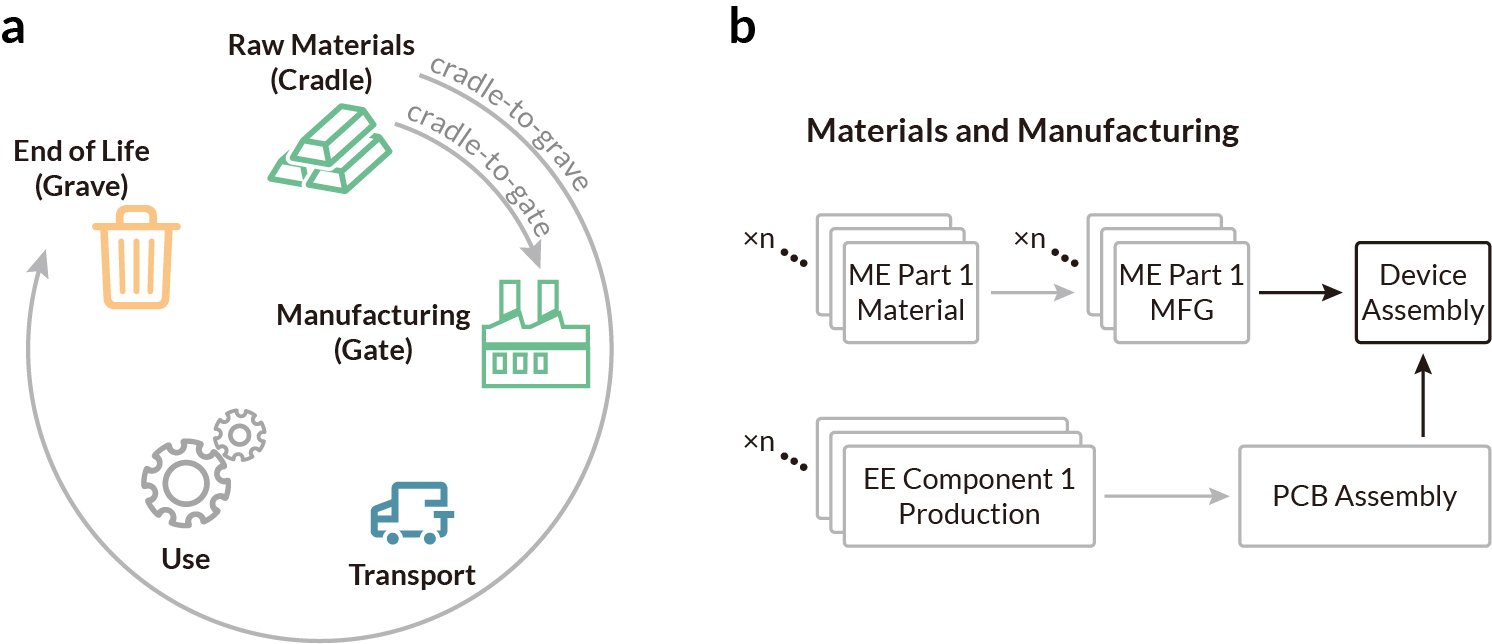}
    \caption{Life cycle assessment system boundaries and methodology for electronic devices. a, Five life cycle stages and example LCA system boundaries. b, Example high-level block diagram of the LCA methodology used by Amazon Devices Sustainability~\cite{amazon-lca-doc}. Diagrams adapted from~\cite{zhang_deltalca_2024}.
    }
    \label{fig:ED_1}
\end{figure}

\begin{figure}[h]
    \centering
    \includegraphics[width=0.6\linewidth]{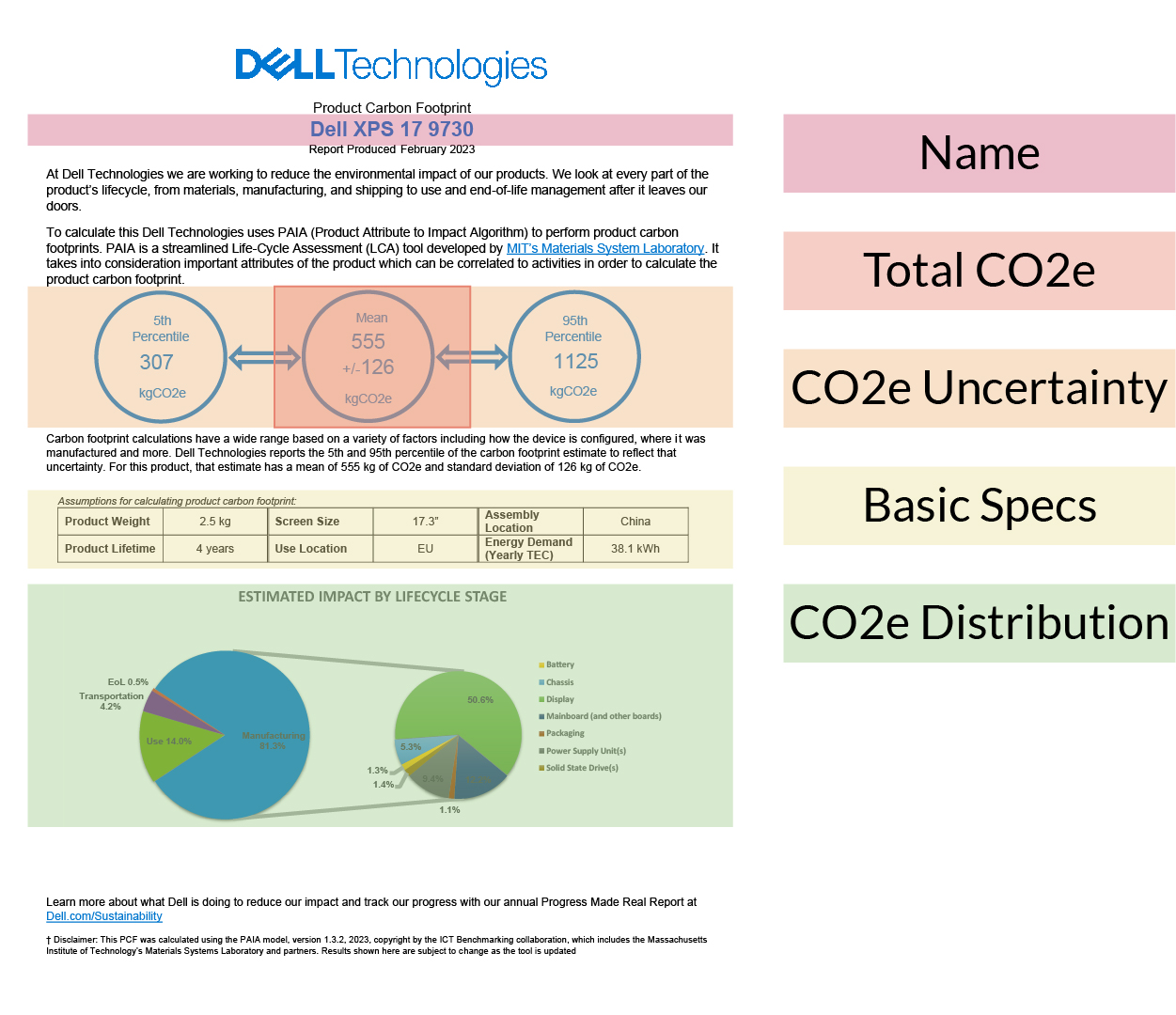}
    \caption{Example product carbon footprint report. Example of a product carbon footprint disclosure from Dell~\cite{dell_xps_17_pcf_2023}. Key information, including the product name, total CO2e, uncertainty, and distributions, is parsed for use as a comparison benchmark.}
    \label{fig:ED_2}
\end{figure}

\clearpage
\begin{figure}[h]
    \centering
    \includegraphics[width=\linewidth]{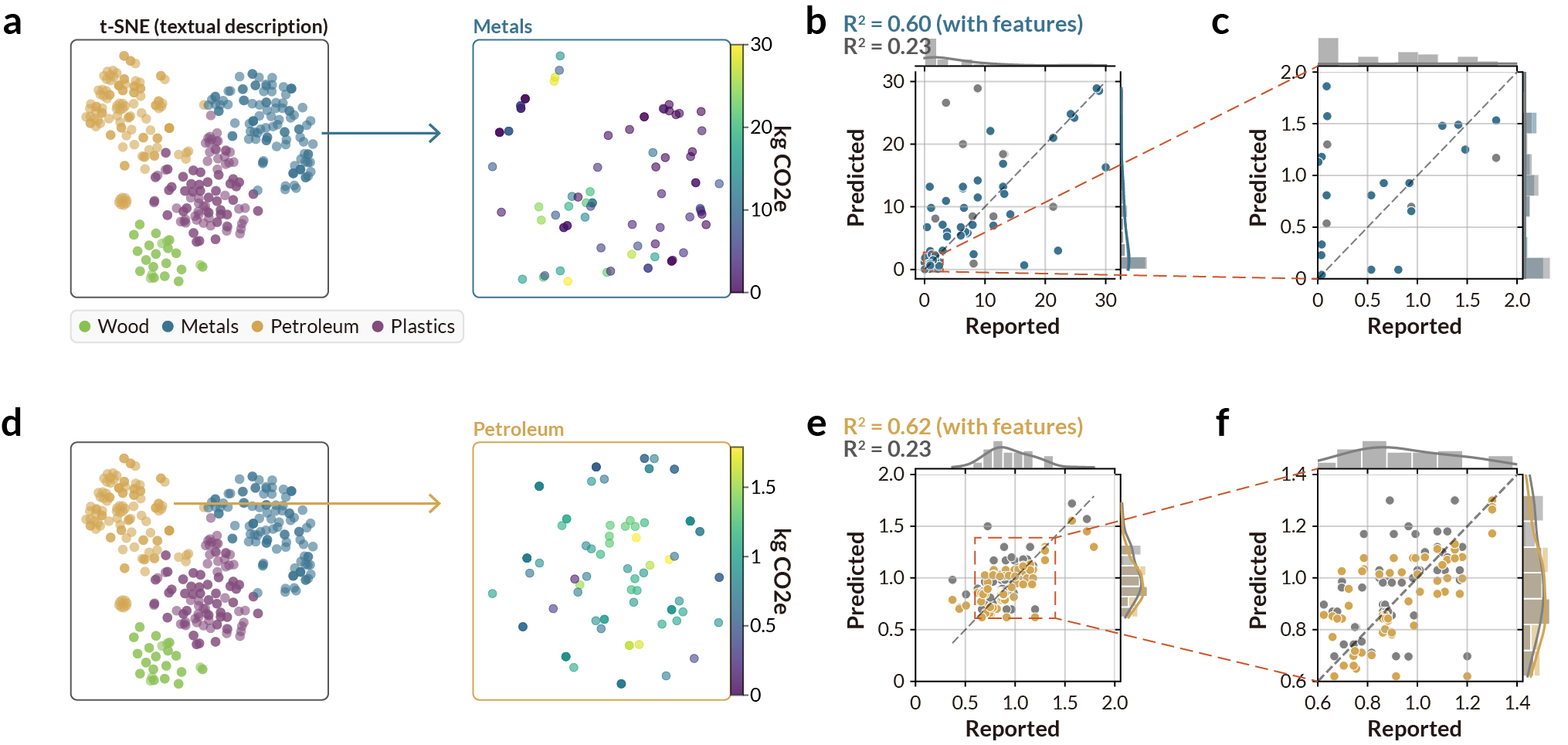}
    \caption{Emission factor estimation for metals and petroleum material classes. a-f, t-SNE plot of raw material entries from ecoinvent based on embeddings of their textual descriptions. Zoom-in on metals (a) and petroleum (d) with points colored by carbon intensity. Each data point represents one LCA database entry. Scatter plot of predicted and carbon intensities from ecoinvent for metals (b,c) and petroleum (e,f), using textual description embeddings alone or combined with domain-specific features (e.g., density); marginal distributions shown.
    }
    \label{fig:ED_raw_materials}
\end{figure}

\clearpage
\begin{figure}[h]
    \centering
    \includegraphics[width=0.75\linewidth]{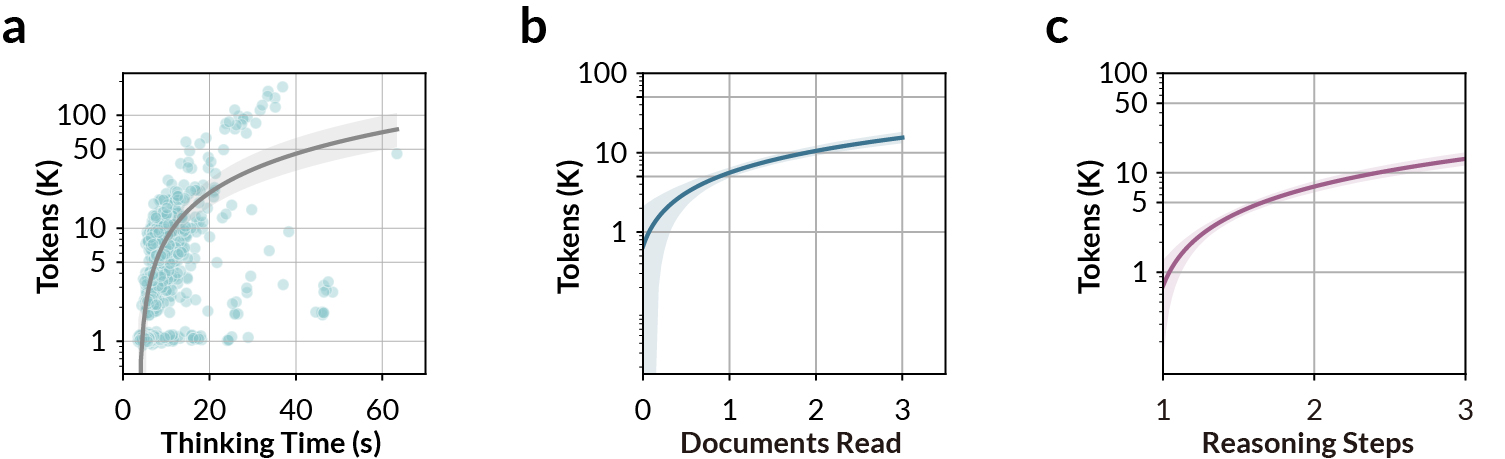}
    \caption{Computational cost of agentic retrieval scaling. 
    a-c, Token usage increases with thinking time (a), number of documents read (b), and reasoning steps (c) during agent self-play.
    All data are N = 200 independent agent runs; shaded regions indicate 95\% CI of regression.
    }
    \label{fig:ED_token}
\end{figure}

\end{document}